\documentclass{ieeeaccess}
\usepackage{cite}
\usepackage{amsmath,amssymb,amsfonts}
\usepackage{algorithmic}
\usepackage{graphicx}
\usepackage{textcomp}

\def\BibTeX{{\rm B\kern-.05em{\sc i\kern-.025em b}\kern-.08em
    T\kern-.1667em\lower.7ex\hbox{E}\kern-.125emX}}
\begin{document}

\doi{}

\title{CIFAR-10 IMAGE CLASSIFICATION USING FEATURE ENSEMBLES}
\author{\uppercase{Felipe O. Giuste}\authorrefmark{1}\textsuperscript{*} and \uppercase{Juan C. Vizcarra}\authorrefmark{2}\textsuperscript{*}}
\address[1]{Biomedical Engineering, Georgia Institute of Technology, Atlanta, GA, USA (e-mail:  fgiuste@gatech.com)}
\address[2]{Biomedical Engineering, Georgia Institute of Technology, Atlanta, GA, USA (e-mail:  jvizcarra3@gatech.edu)}

\corresp{\textsuperscript{*} The authors contributed equally to this work.}

\begin{abstract}
Image classification requires the generation of features capable of detecting image patterns informative of group identity. The objective of this study was to classify images from the public CIFAR-10 image dataset by leveraging combinations of disparate image feature sources from both manual and deep learning approaches. Histogram of oriented gradients (HOG) and pixel intensities successfully inform classification (53\% and 59\% classification accuracy, respectively), yet there is much room for improvement. VGG16 with ImageNet trained weights and a CIFAR-10 optimized model (CIFAR-VGG) further improve upon image classification (60\% and 93.43\% accuracy, respectively). We further improved classification by utilizing transfer learning to re-establish optimal network weights for VGG16 (TL-VGG) and Inception ResNet v2 (TL-Inception) resulting in significant performance increases (85\% and 90.74\%, respectively), yet fail to surpass CIFAR-VGG. We hypothesized that if each generated feature set obtained some unique insight into the classification problem, then combining these features would result in greater classification accuracy, surpassing that of CIFAR-VGG. Upon selection of the top 1000 principal components from TL-VGG, TL-Inception, HOG, pixel intensities, and CIFAR-VGG, we achieved testing accuracy of 94.6\%, lending support to our hypothesis.
\end{abstract}


\titlepgskip=-15pt

\maketitle

\section{INTRODUCTION}
\label{sec:introduction}
\PARstart{T}{he} CIFAR-10 dataset includes 10 classes, with 5 thousand training images per class and 1 thousand testing images per class \cite{cifar}. The images are of dimensionality 32x32 with three color channels [Fig 1].  State-of-the-art methods with the highest classification accuracies involve giant neural networks and custom architectures which reach the limits of performance (greater than 98\% accuracy [table 1]) \cite{2}\cite{6}. These methods involve significant customization and training times, and may not be robust to generalized image classification outside of CIFAR-10 labels. CIFAR-VGG is a modified VGG16 architecture with additional dropout and weight decay to reduce overfitting potential \cite{3}\cite{4}. These changes lead to competitive published classification accuracy of 93.43\% on the CIFAR-10 testing dataset. We used this network architecture as our gold standard because it is a simplified network compared to modern, highly-specific and manually-optimized, architectures. Given the wide variety of available solutions to image classification problems, we were curious to see if different convolutional neural network architectures learned unique features, and if combining these features with histogram of oriented gradients (HOG) features and pixel intensity values would result in improved predictive power. In lieu of generating a single optimized architecture to optimize classification accuracy, we instead chose to integrate features sets obtained from HOG, pixel intensities, VGG16 and Inception ResNet v2 with transfer learning (TL-VGG and TL-Inception respectively), and CIFAR-VGG to obtain image classification accuracies that surpass CIFAR-VGG alone.

 \Figure[t](topskip=0pt, botskip=0pt, midskip=0pt){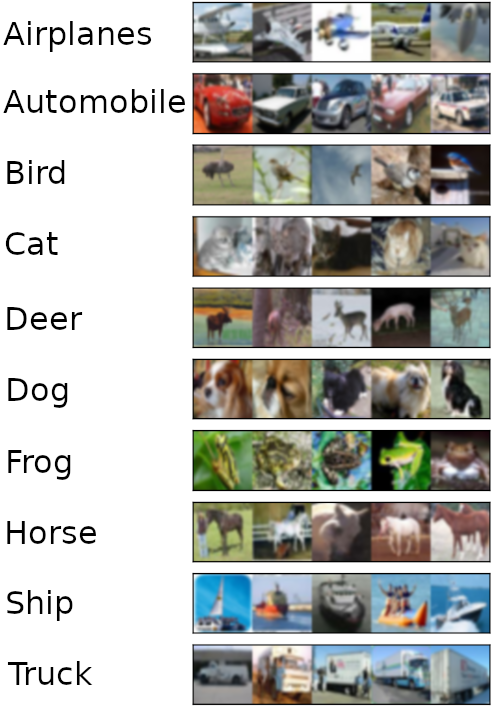}
 {\textbf{Images from the CIFAR-10 dataset}. The dataset contains images grouped into 10 unique classes (rows). Each class contains a subset of images belonging to the training dataset (5 thousand) and testing/validation dataset (1 thousand images). Images are of small resolution (32x32) RGB color images.\label{fig1}}

\section{BACKGROUND}
\label{sec:background}
Convolutional Neural Networks (CNN) optimize convolution kernels which map image characteristics to lower dimensionality, allowing for improved memory management while maintaining relevant image information. Fully connected layers within CNNs allows optimized weighing of convolutional features, and eventual mapping onto discrete categories. 

Data augmentation increases training dataset size by adding images derived from the original dataset via simple transformations, such as horizontal and vertical flipping, small angle rotation, and translation. These new images should retain their original class membership without loss of meaningful group information. These methods may also improve model generalizability. 

Ensemble methods are an approach to algorithm improvement which seeks to combine the results of relatively weak classifiers with the assumption that significant learned features and transformations are retained, whereas inconsistent and unhelpful perspectives are reduced in the final model \cite{5}. This requires that the original weak classifiers are sufficiently independent such that their combination results in novel strategies. We developed and trained fully-connected neural networks (FCNN) which receives pre-computed features from each image and assigns labels according to internal weights which minimize error across the training dataset. The features used include: HOG, pixel intensities, TL-VGG, TL-Inception, and CIFAR-VGG final dense layer weights [Fig. 2]. Training dataset was augmented with horizontally flipped training images to increase the size from 5K to 10K images per class. All Deep Learning (DL) was implemented with TensorFlow version 2 and Keras in Python. Testing dataset was used to calculate validation accuracy throughout training. Final model accuracy was taken to be the validation accuracy upon completion of model training. All code is available on Github (\underline{https://github.com/jvizcar/feature\_ensembles\_clf}).

\subsection{HOG Feature Generation}
HOG is a common image feature extraction method that partitions the image into equally sized sections and calculates the intensity gradient direction within each section. A histogram of these orientations are then created and the

\begin{table}[h]
\caption{\textbf{Top CIFAR-10 Classifiers} \cite{6}}
\begin{tabular}{p{0.85\textwidth}p{0.1\textwidth}}
\textbf{Method}                                                                                                                                                       & \textbf{Accuracy}            \\ \hline
\multicolumn{1}{|l|}{\begin{tabular}[c]{@{}l@{}}GPipe: Efficient Training of Giant Neural Networks using Pipeline Parallelism \cite{1} \end{tabular}} & \multicolumn{1}{l|}{99.00\%} \\ \hline
\multicolumn{1}{|l|}{\begin{tabular}[c]{@{}l@{}}EfficientNet: Rethinking Model Scaling for Convolutional Neural Networks \cite{7} \end{tabular}}      & \multicolumn{1}{l|}{98.90\%} \\ \hline
\multicolumn{1}{|l|}{\begin{tabular}[c]{@{}l@{}}A Survey on Neural Architecture Search \cite{2} \end{tabular}}                                        & \multicolumn{1}{l|}{98.67\%} \\ \hline
\multicolumn{1}{|l|}{AutoAugment: Learning Augmentation Policies from Data \cite{8}}                                                                    & \multicolumn{1}{l|}{98.52\%} \\ \hline
\multicolumn{1}{|l|}{XNAS: Neural Architecture Search with Expert Advice \cite{9}}                                                                      & \multicolumn{1}{l|}{98.40\%} \\ \hline
\end{tabular}
\end{table}

 counts are used as the feature values. The "hog" function in Python’s scikit-image package was used to generate HOG features. We chose this feature set generation method because it provides meaningful information about edge content.

\subsection{Image Pixel Values as Features}
The 32x32, 3-channel, images were flattened into an array of 3072 pixel and channel intensity values. These features represent the entire image, and therefore can be used as raw representations of the original image when input to FCNN.

\subsection{TL-VGG Network Optimization \& Features}
VGG16 initialized with ImageNet weights was trained on the augmented CIFAR-10 training set to optimize network weights \cite{10}. During training, early stopping was implemented using a “patience” parameter of 10 epochs. This means that stopping ensued after 10 consecutive epochs of validation accuracy not increasing. The parameter “min\_delta” was kept at the default value of 0, meaning that any validation accuracy increase would reset the early stopping epoch counter. This limits the effects of overfitting by ensuring that validation accuracy consistently increases. The final softmax layer, mapping to 1000 output classes, was replaced with a dense layer, 50\% dropout layer, and softmax layer mapping to the CIFAR-10 labels [Fig. 3]. These layers were introduced to maximize TL-VGG classification accuracy during the transfer learning process. To extract model features, the top dense layer was removed, along with dropout layer, and the output of the previous dense layer was used as a feature vector resulting in a 512-length labeled feature vector per image.

\subsection{TL-Inception Network Optimization \& Features}
Inception ResNet v2 was retrained on the augmented training set using ImageNet starting weights, after replacement of its top softmax layer with a softmax layer mapping to the CIFAR labels, to generate the TL-Inception model. Features were generated by removing the top softmax and dropout layers of the network. The final feature set included 1024 features representing each image. 

\subsection{CIFAR-VGG Features}
CIFAR-VGG features were obtained by removing the top softmax and dropout layers to generate 512 image features for every CIFAR-10 image [Fig. 4]. The trained model weights were loaded directly from provided source files (\underline{https://github.com/geifmany/cifar-vgg}).
\\
\\
\\
\\
\\
\\
\\

\Figure[t!](topskip=0pt, botskip=0pt, midskip=0pt){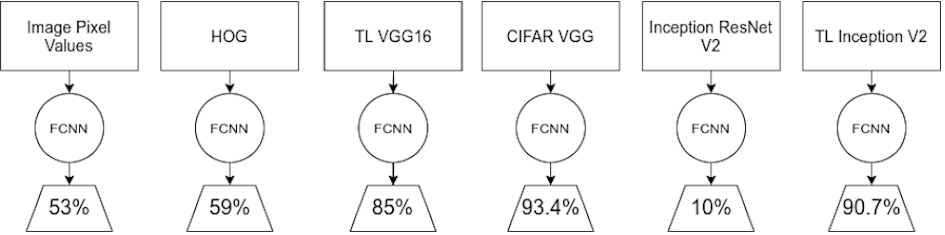}
{\textbf{FCNN single feature set performance.} Six methods were used to generate image features, four of which utilize deep learning to extract features. Image features sets were then used to train a FCNN architecture and testing data used to generate final accuracy estimates. Top boxes represent the individual feature set generation method, bottom polygon shows the accuracy obtained by the trained FCNN.\label{fig2}}

\Figure[t!](topskip=0pt, botskip=0pt, midskip=0pt){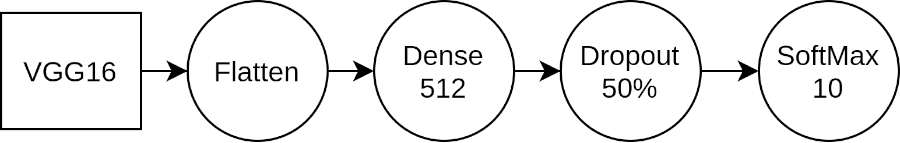}
{\textbf{Modification of VGG16 used with transfer learning.} VGG16 architecture model was modified by adding a flattened layer, followed by a dense layer with dropout and softmax for 10 classes. Transfer learning was used during training of the model with early stopping.\label{fig3}}

\Figure[t!](topskip=0pt, botskip=0pt, midskip=0pt){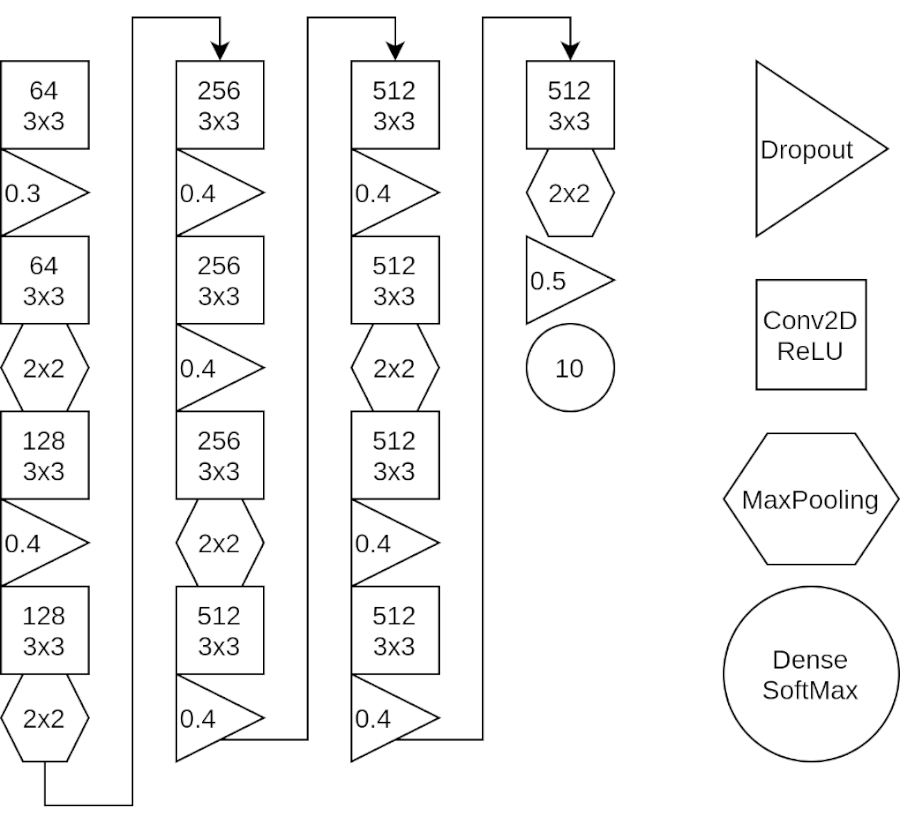}
{\textbf{CIFAR-VGG model.} Architecture of the CIFAR-VGG model adapted from (https://github.com/geifmany/cifar-vgghttps://github.com/geifmany/cifar-vgg). Features were extracted from this previously trained model.\label{fig4}}

\subsection{FCNN Generation and Training}
In order to test the ability of each feature set to categories test set images, a simple architecture FCNN was developed. Features are connected with a dense layer of 300 neurons, followed by 50\% dropout layer, and 100 neuron dense layer fully connected with a softmax layer leading to 10 classes. Dropout and early stopping (patience of 10 epochs was used) were included to prevent overfitting of the network. This was especially of concern because the CNN features were already the result of significant training.
\\
\\
\\

\section{EXPERIMENTS}
\label{sec:experiments}
\subsection{Baseline Model Accuracy}
Baseline CNN testing image classification accuracy was measured on VGG16 (ImageNet weights), TL-VGG, CIFAR-VGG, Inception ResNet v2 (ImageNet weights), and TL-Inception CNNs. It was observed that Inception ResNet v2 had chance performance (10\% accuracy) on the 10-class problem, potentially signifying a lack of generalizability of the pretrained model or a lack of similarity between the CIFAR-10 dataset and ImageNet. Both are suspected to be true given the relatively large network size of Inception ResNet v2 as well as the significant image dimension differences of ImageNet (128x128, 3 channels) and CIFAR-10 (32x32, 3 channels).

The two re-trained models with transfer learning, TL-VGG and TL-Inception, significantly outperformed their original networks, with a 15\% increase in classification accuracy difference between VGG architectures (60\% to 85\% accuracy), and a 75\% increase in TL-Inception relative to ImageNet weighted architecture (10\% to 85\% accuracy). CIFAR-VGG outperformed all CNN models with a 93.43\% test-set classification accuracy. 

\subsection{Individual Feature Set Classification Success}
To compare feature set classification potential, HOG, pixel intensities, TL-VGG, TL-Inception, and CIFAR-VGG feature sets were used to independently train an FCNN and the resulting network was used to calculate testing accuracy [Fig. 2]. We observed modest performance of HOG and pixel intensities features sets (59\% and 53\% accuracy, respectively), good performance with TL-VGG and TL-Inception features sets (85\% and 90.74\% accuracy, respectively), and excellent performance with CIFAR-VGG feature set (93.43\%). Although CIFAR-VGG features did not improve with further training on an FCNN, it still outperformed the others.

\Figure[h!](topskip=0pt, botskip=0pt, midskip=0pt){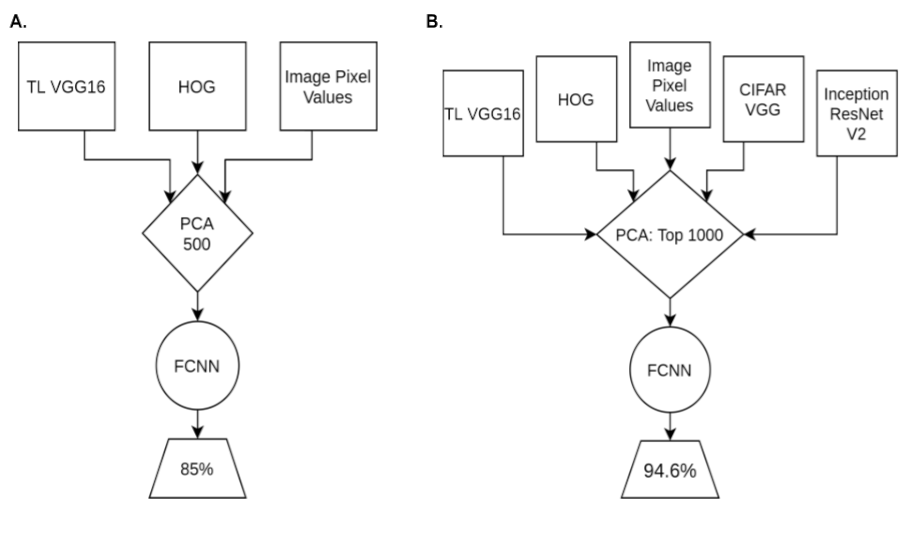}
{\textbf{Ensemble features results.} (a) Combining features from VGG16 TL model, HOG, and pixel intensities and taking the top 500 PCA resulted in an 85\% accuracy when using the features to train a new FCNN. (b) Best performance was obtained when combining the 5 feature sets and taking the top 1000 PCA and training a new FCNN with these features (94.6\% accuracy)\label{fig5}}

\Figure[h!](topskip=0pt, botskip=0pt, midskip=0pt){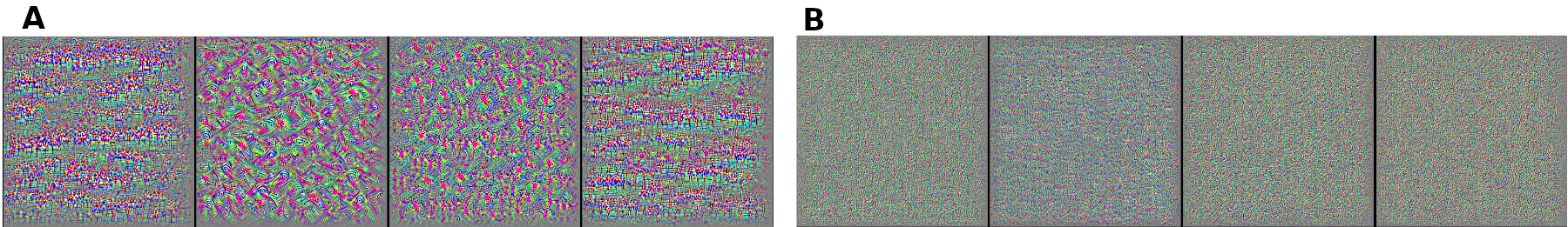}
{\textbf{(A)} TL-VGG and \textbf{(B)} CIFAR-VGG features from final convolutional layer. These images are representative of their corresponding layers.\label{fig6}}

\subsection{Feature Set Ensembles}
To test our hypothesis that combining disparate features sets would lead to improved results, we trained 3 separate FCNNs. The first FCNN trained with the top 500 principal components (after principal component analysis (PCA)) of TL-VGG, HOG, and pixel intensity feature sets (95\% explained variance) to achieve 85\% test-set accuracy (no improvement from TL-VGG alone). The second FCNN trained on TL-VGG and TL-Inception features directly (512+1024 features) to achieve 91.12\% accuracy; this was a significant improvement from individual model performances. The final model trained with the top 1000 principal components of TL-VGG, HOG, pixel intensity, CIFAR-VGG, and TL-Inception feature sets. This final model obtained the best performance with 94.6\% classification accuracy on the testing dataset, surpassing the benchmark established by CIFAR VGG (93.43\%) by a small margin [Fig. 5]. 

\section{CONCLUSION}
We obtained notable performance improvement with an ensemble of feature sets compared to individual classification models. Success was measured by CIFAR-10 test set classification accuracy (\%). We believe that this lends some support for the hypothesis that each feature set represents unique facets of variability between class labels, and that by merging a diverse array of successful features sets, a more complete description of relevant differences emerges to aid in image classification. This hypothesis is further supported by the drastic differences between CNN model features upon visual inspection [Fig. 6].  It appears that TL-VGG features detect larger image patterns as compared with CIFAR-VGG. The differences in CNN features suggests that there are many ways of extracting meaningful features from these images, and that combining them may improve classification. Non-CNN features are known to be very different from CNN features as they describe likely meaningful patterns found in the images (eg. intensity gradients in HOG). The addition of non-CNN features further enhances the diversity of image descriptions available for classification by the FCNN, allowing for more robust image classification relative to single feature set training alone.

Error analysis shows that although 5-Set PCA FCNN model exceeded the others in overall image classification accuracy, it had a more difficult time differentiating between cats and dogs relative to CIFAR-VGG [Fig. 7]. The confusion matrix also showed a decrease in airplane misclassification which suggests that the inclusion of additional independent feature sets in the final model improves the classification of high-error (relatively speaking) labels. Correctly classified images had high confidence for the correct label, and misclassified images seemed to contain the correct label within the top three class results [Fig. 8]. Inspected images that were incorrectly labeled were difficult to make out due to high background noise, low light conditions, and ambiguous class membership (automobile vs. truck). The low resolution of images increased the ambiguity of many images, yet the final model which utilized a feature ensemble was robust enough to consistently distinguish between distinct labels. In general, the inclusion of multiple features sets resulted in improved classification accuracy, and failure in situations which made it difficult for humans to classify as well. 

Although the generation of multiple feature sets takes additional time and computational power, the potential for performance increases may justify these costs in some scenarios. These results may also be applicable in other classification tasks, especially those including a small margin for error (eg. clinical decision support or automated driving). This work emphasizes the need for continued exploration of improved feature generation methods to maximize the utility of current deep learning architectures. 

\bibliography{access} 

\begin{thebibliography}{10}
\providecommand{\url}[1]{#1}
\csname url@samestyle\endcsname
\providecommand{\newblock}{\relax}
\providecommand{\bibinfo}[2]{#2}
\providecommand{\BIBentrySTDinterwordspacing}{\spaceskip=0pt\relax}
\providecommand{\BIBentryALTinterwordstretchfactor}{4}
\providecommand{\BIBentryALTinterwordspacing}{\spaceskip=\fontdimen2\font plus
\BIBentryALTinterwordstretchfactor\fontdimen3\font minus
  \fontdimen4\font\relax}
\providecommand{\BIBforeignlanguage}[2]{{%
\expandafter\ifx\csname l@#1\endcsname\relax
\typeout{** WARNING: IEEEtran.bst: No hyphenation pattern has been}%
\typeout{** loaded for the language `#1'. Using the pattern for}%
\typeout{** the default language instead.}%
\else
\language=\csname l@#1\endcsname
\fi
#2}}
\providecommand{\BIBdecl}{\relax}
\BIBdecl

\bibitem{cifar}
A.~Krizhevsky, I.~Sutskever, and G.~E. Hinton, ``{ImageNet} classification with
  deep convolutional neural networks,'' in \emph{Advances in Neural Information
  Processing Systems 25}, F.~Pereira, C.~J.~C. Burges, L.~Bottou, and K.~Q.
  Weinberger, Eds.\hskip 1em plus 0.5em minus 0.4em\relax Curran Associates,
  Inc., 2012, pp. 1097--1105.

\bibitem{2}
M.~Wistuba, A.~Rawat, and T.~Pedapati, ``A survey on neural architecture
  search,'' May 2019.

\bibitem{6}
``{CIFAR-10} on {Benchmarks.AI},'' \url{https://benchmarks.ai/cifar-10},
  accessed: 2020-2-5.

\bibitem{3}
K.~Simonyan and A.~Zisserman, ``Very deep convolutional networks for
  {Large-Scale} image recognition,'' Sep. 2014.

\bibitem{4}
S.~Liu and W.~Deng, ``Very deep convolutional neural network based image
  classification using small training sample size,'' in \emph{2015 3rd {IAPR}
  Asian Conference on Pattern Recognition ({ACPR})}, Nov. 2015, pp. 730--734.

\bibitem{5}
Y.~Chen, Y.~Yang, W.~Wang, and C.~C.~Jay~Kuo, ``Ensembles of
  feedforward-designed convolutional neural networks,'' Jan. 2019.

\bibitem{1}
Y.~Huang, Y.~Cheng, A.~Bapna, O.~Firat, M.~X. Chen, D.~Chen, H.~Lee, J.~Ngiam,
  Q.~V. Le, Y.~Wu, and Z.~Chen, ``{GPipe}: Efficient training of giant neural
  networks using pipeline parallelism,'' Nov. 2018.

\bibitem{7}
M.~Tan and Q.~V. Le, ``{EfficientNet}: Rethinking model scaling for
  convolutional neural networks,'' May 2019.

\bibitem{8}
E.~D. Cubuk, B.~Zoph, D.~Mane, V.~Vasudevan, and Q.~V. Le, ``{AutoAugment}:
  Learning augmentation policies from data,'' May 2018.

\bibitem{9}
N.~Nayman, A.~Noy, T.~Ridnik, I.~Friedman, R.~Jin, and L.~Zelnik-Manor,
  ``{XNAS}: Neural architecture search with expert advice,'' Jun. 2019.

\bibitem{10}
J.~Deng, W.~Dong, R.~Socher, L.~Li, K.~Li, and L.~Fei-Fei, ``{ImageNet}: A
  large-scale hierarchical image database,'' in \emph{2009 {IEEE} Conference on
  Computer Vision and Pattern Recognition}, Jun. 2009, pp. 248--255.

\end{thebibliography}
\bibliographystyle{IEEEtrans}

\onecolumn

\Figure[ht!](topskip=0pt, botskip=0pt, midskip=0pt){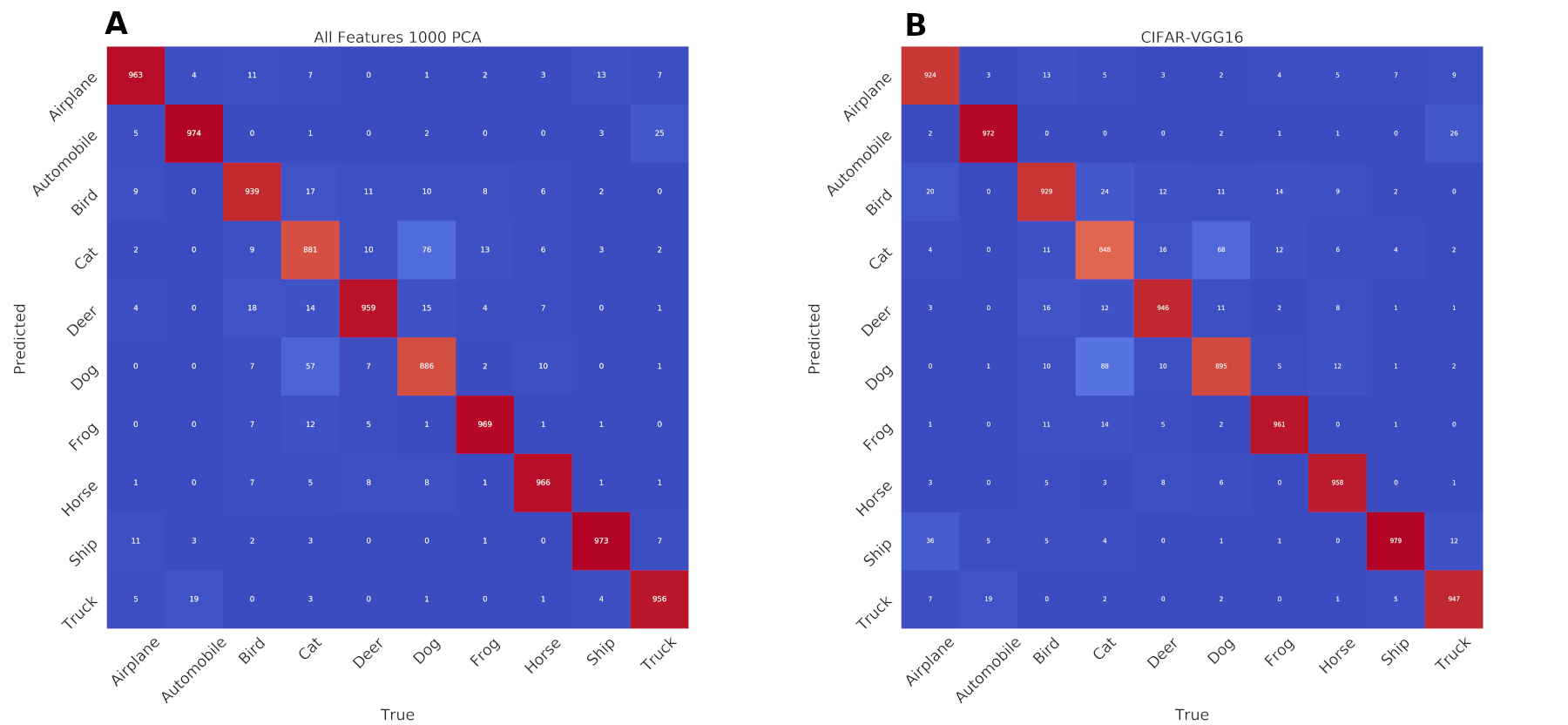}
{\textbf{Confusion Matrices of leading models} \textbf{(A)} Top 1000 PCA using all 5 feature sets. \textbf{(B)} CIFAR-VGG16 model.\label{fig7}}

\Figure[ht!](topskip=0pt, botskip=0pt, midskip=0pt){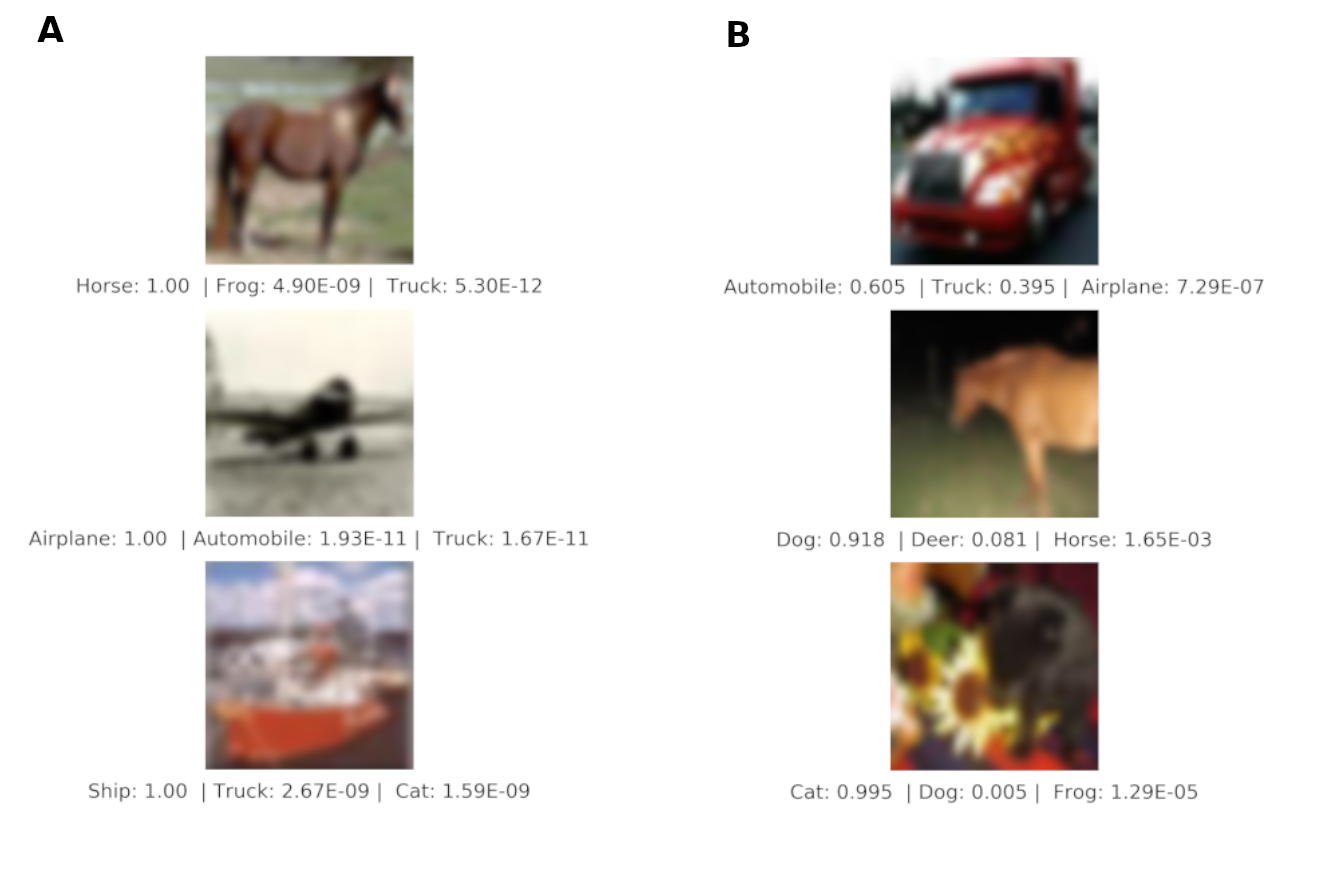}
{\textbf{5-Set PCA Error Analysis.} Correct \textbf{(A)} and Incorrect \textbf{(B)} 5-Set PCA model labeled images with top three labels shown.\label{fig8}}

\EOD

\end{document}